\documentclass[10pt,journal,compsoc,]{IEEEtran}
\usepackage{amssymb}
\usepackage{amsmath}
\usepackage{booktabs} 
\usepackage[table]{xcolor}
\usepackage{multirow}
\newcommand\sysName{mmEgoHand}

\usepackage{orcidlink}

\ifCLASSINFOpdf
\else
\fi
\begin{document}

\title{\sysName: Egocentric Hand Pose Estimation and Gesture Recognition with  Head-mounted Millimeter-wave Radar and IMU}

\author{Yizhe~Lv\orcidlink{0009-0005-3319-7561},~
        Tingting~Zhang\orcidlink{0009-0007-2157-4360},~
        Zhijian~Wang\orcidlink{0009-0009-8143-2860},~
        Yunpeng~Song\orcidlink{0000-0002-4186-0408},~\\
        Han~Ding\orcidlink{0000-0002-5274-7988},~\IEEEmembership{Senior~Member,~IEEE},~
        Jinsong Han\orcidlink{0000-0001-5064-1955},~\IEEEmembership{Senior~Member,~IEEE},~
        Fei Wang\orcidlink{0000-0002-0750-6990}
\thanks{Under Review.}
\thanks{Yizhe Lv~(email:  lvyizhe@stu.xjtu.edu.cn), Tingting Zhang~(email: zhang\_tt@lvyizhe@stu.xjtu.edu.cn), Zhijian Wang~(email: wangzhijian@stu.xjtu.edu.cn), and Fei Wang~(email: feynmanw@xjtu.edu.cn) are with the School of Software Engineering, Xi'an Jiaotong University, Xi'an 710049, China.}
\thanks{Yunpeng Song~(email: yunpengs@xjtu.edu.cn) is with School of Cyber Science and Engineering, Xi'an Jiaotong University, Xi'an 710049, China.}
\thanks{Han Ding~(email: dinghan@xjtu.edu.cn) is with the School of Computer Science and Technology, Xi'an Jiaotong University, Xi'an 710049, China. }
\thanks{Jinsong Han~(email: hanjinsong@xjtu.edu.cn) is with the College of Computer Science and Technology, Zhejiang University, Hangzhou 310027, China.}
\thanks{Fei Wang is the corresponding author.}
}

\markboth{Journal of \LaTeX\ Class Files,~Vol.~00, No.~0, January~2025}%
{\sysName: Egocentric Hand Pose Estimation and Gesture Recognition with  Head-mounted Millimeter-wave Radar and IMU}

\maketitle

\begin{abstract}
Recent advancements in millimeter-wave (mmWave) radar have demonstrated its potential for human action recognition and pose estimation, offering privacy-preserving advantages over conventional cameras while maintaining occlusion robustness, with promising applications in human-computer interaction and wellness care. However, existing mmWave systems typically employ fixed-position configurations, restricting user mobility to predefined zones and limiting practical deployment scenarios. We introduce \sysName, a head-mounted egocentric system for hand pose estimation to support applications such as gesture recognition, VR interaction, skill digitization and assessment, and robotic teleoperation. \sysName~ synergistically integrates mmWave radar with inertial measurement units (IMUs) to enable dynamic perception. The IMUs actively compensate for radar interference induced by head movements, while our novel end-to-end Transformer architecture simultaneously estimates 3D hand keypoint coordinates through multi-modal sensor fusion. This dual-modality framework achieves spatial-temporal alignment of mmWave heatmaps with IMUs, overcoming viewpoint instability inherent in egocentric sensing scenarios. We further demonstrate that intermediate hand pose representations substantially improve performance in downstream task, e.g., VR gesture recognition. Extensive evaluations with 10 subjects performing 8 gestures across 3 distinct postures- standing, sitting, lying - achieve 90.8\% recognition accuracy, outperforming state-of-the-art solutions by a large margin. Dataset and code are available at \url{https://github.com/WhisperYi/mmVR}.
\end{abstract}

\begin{IEEEkeywords}
Human sensing, Hand pose estimation, Gesture recognition, Millimeter-wave radar, IMUs, Human-computer interaction
\end{IEEEkeywords}

\section{Introduction}\label{sec:introduction}

\begin{figure}[t]
    \centering
    \includegraphics[width=1\linewidth]{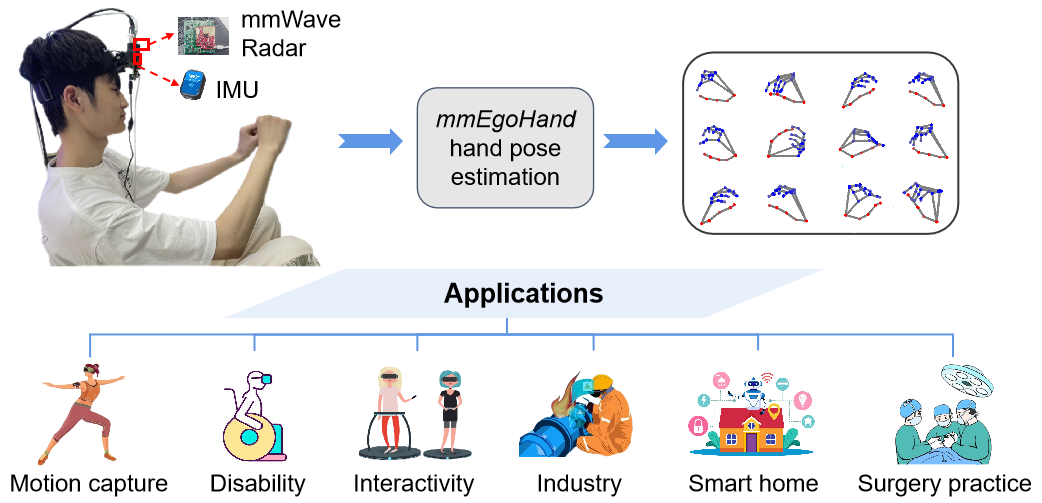}
     \caption{We present \sysName, an interaction gesture recognition system utilizing head-mounted millimeter-wave radar and IMUs as an alternative or complement to downward-facing cameras for enhancing personal privacy.}
  \label{fig:hm_introduction}
\end{figure}

Human activity recognition and pose estimation are foundational technologies that support a wide range of applications, including human behavior analysis, wellness care, human-computer interaction, and gaming. Traditionally, these tasks have relied on camera-based solutions~\cite{feichtenhofer2019slowfast,cao2017realtime}. More recently, wireless sensing alternatives such as Wi-Fi~\cite{wang2019person} and custom-built radar systems~\cite{zhao2018through} have emerged to address the limitations of visual methods, offering improved privacy and robustness to occlusion. However, Wi-Fi-based approaches are sensitive to environmental variations and subject orientation~\cite{wang2025survey,chen2023cross}, while custom radar systems face challenges in adoption due to hardware accessibility. These constraints have spurred interest in commercially available millimeter-wave (mmWave) radar systems, such as the Texas Instruments (TI) IWR/AWR series, which offer a compelling trade-off between deployment practicality and high-resolution motion sensing for human activity recognition and pose estimation.

The evolution of mmWave radar-based human sensing has progressed from coarse activity recognition~\cite{liu2020real,wang2021m,liu2022mtranssee,xie2022mmfit,liu2024real,deng2023midas++,zhao2025federated} to fine-grained pose estimation~\cite{xue2021mmmesh,zhang2022synthesized,xue2023towards,xue2022m4esh,chen2022mmbody,kong2022m3track,lee2023hupr,ding2023mi,zhang2024super}. Most existing systems employ frontal, static deployments, requiring users to stay within a constrained sensing volume—an approach that limits applicability in dynamic environments. Recent advances in egocentric sensing, such as mmEgo~\cite{li2023egocentric} and Argus~\cite{duan2024argus}, have begun to overcome this limitation by leveraging head-mounted radar to capture body-reflected signals for pose estimation. These systems enable full-body tracking while preserving human movement, shifting mmWave sensing from environment-anchored perception to mobile, user-centric motion capture.

In this work, we introduce \sysName, an egocentric hand pose estimation system. While prior mmWave methods primarily focus on full-body tracking, our system targets fine-grained hand articulation to support applications such as gesture recognition, VR interaction, skill digitization and assessment, and robotic teleoperation. The design of \sysName~ is guided by three key considerations: (1) Existing frontal-view systems~\cite{dong2024mmhand,kong2024mmhand,liu2022leveraging} typically estimate single-hand poses, whereas many real-world gestures involve both hands. We thus design a unified framework capable of handling both single- and two-hand configurations. (2) Head-mounted radar introduces ego-motion artifacts, as head movements dynamically alter the radar’s perspective, changing signal characteristics for the same hand gesture. Our system incorporates active motion compensation to disentangle hand motion from head-induced signal variations. (3) Traditional motion capture systems, such as VICON~\cite{vicon}, are cumbersome for annotating hand poses. We leverage a lightweight annotation pipeline that ensures high-quality ground truth while significantly simplifying deployment, improving adaptability across application scenarios.

To address the first consideration, we adapt an end-to-end Wi-Fi-based multi-person pose estimation~\cite{yan2024person}. In the original design, the system automatically detects a variable number of individuals and estimates their body keypoints. Similarly, \sysName~ is designed to output a variable number of hands with a set-based Hungarian matching algorithm~\cite{kuhn1955hungarian} and estimate keypoints for each detected hand in an end-to-end manner. This modification enables \sysName~ to support both one-handed and two-handed interaction tasks, making it more versatile and applicable to a wider range of scenarios compared to existing methods such as mm4Arm~\cite{liu2022leveraging} and mmHand~\cite{kong2024mmhand,dong2024mmhand}, which only support fixed one-handed interactions. Furthermore, we augment the original architecture with a context decoder that takes a sequence of mmWave heatmaps as input, leveraging temporal information to improve the accuracy of hand pose estimation. 

For the second consideration, we draw inspiration from the hardware setup of mmEgo~\cite{li2023egocentric}, attaching an inertial measurement unit (IMU) near the mmWave radar to capture head motion. The IMU data is temporally synchronized with the mmWave input and fed jointly into the network. By fusing these two modalities, \sysName~ can compensate for ego-motion artifacts caused by head movement, leading to more stable and accurate hand pose estimation. To address the third consideration, we employ a lightweight annotation strategy during data collection. A standard web camera is placed next to the subject to record hand movements, and hand keypoints are automatically extracted from the video using Google MediaPipe Hand Landmark SDK. We then perform extensive manual checking and filtering, removing keypoints with poor tracking quality, particularly under fast hand motion conditions, and take the remaining as ground truth annotations for training and evaluation.
Similar camera-based automatic annotation schemes have become increasingly popular in prior work~\cite{wang2019person,zhao2018through,duan2024argus}, offering a practical trade-off between annotation quality and deployment simplicity.

To evaluate \sysName, we recruited 10 volunteers who performed hand movements in three different scenes and under three distinct postures: standing, sitting, and lying. \sysName~ achieved a mean per-joint position error (MPJPE) of 72.73mm for hand pose estimation. In comparison, using mmWave radar alone resulted in an MPJPE of 96.42mm, highlighting the effectiveness of incorporating IMU data. Furthermore, we used the estimated hand poses from \sysName~ as intermediate representations for a downstream hand gesture recognition task. \sysName~ achieved an accuracy of 90.80\%, substantially outperforming state-of-the-art approaches such as mGesNet~\cite{liu2020real} (81.34\%), mSeeNet~\cite{liu2022mtranssee} (84.60\%), and mmGesture~\cite{yan2023mmgesture} (68.66\%). These results highlight that \sysName~ not only enhances hand pose estimation, but also provides more reliable and informative features for downstream applications.
Our contributions are summarized as follows:

(1) We introduce mmEgoHand, the first egocentric system capable of human hand pose estimation and gesture recognition (see Table~\ref{tab:existing-work} for a comparison with prior work).

(2) We propose a novel dual-decoder architecture that simultaneously addresses spatial information and temporal coherence, yielding significant improvements over
state-of-the-art approaches

(3) We collect and open-source a large-scale dataset of VR interaction gestures captured using head-mounted mmWave radar and IMU, totaling 26GB, to support future research in human sensing. Dataset and code are available at \url{https://github.com/WhisperYi/mmVR}

\begin{table}[t]
\caption{\sysName~ is the first egocentric system capable of simultaneously capturing both one-hand and two-hand poses using mmWave radar.}
\centering
\setlength{\tabcolsep}{3pt} 
\begin{tabular}{lccc}
\toprule
\textbf{Solutions}                                      & \textbf{Radar}                                                & \textbf{Sensing View} & \textbf{Body Part} \\ \midrule
mmMesh~\cite{xue2021mmmesh}           & AWR1843            & Frontal               & Body               \\
SynMotion~\cite{zhang2022synthesized} & IWR1443             & Frontal               & Body               \\
mmGPE~\cite{xue2023towards}           & AWR1843             & Frontal               & Body               \\
M$^4$esh~\cite{xue2022m4esh}          & AWR1843             & Frontal               & Body               \\
mmBody~\cite{chen2022mmbody}          & Arbe Phoenix           & Frontal               & Body               \\
m$^3$Track~\cite{kong2022m3track}     & IWR1443             & Frontal               & Body               \\
HUPR~\cite{lee2023hupr}               & IWR1843            & Frontal               & Body               \\
Mi-Mesh~\cite{ding2023mi}               & AWR1443           & Frontal               & Body               \\
SUPER~\cite{zhang2024super}           & IWR6843           & Frontal               & Upper              \\
mm4Arm~\cite{liu2022leveraging}           & IWR6843           & Frontal               & \textbf{One hand}              \\
mmHand~\cite{kong2024mmhand}           & AWR1443            & Frontal               & \textbf{One hand}              \\
mmHand~\cite{dong2024mmhand}           & IWR1843            & Frontal               & \textbf{One hand}              \\
mmEgo~\cite{li2023egocentric}         & IWR6843            & \textbf{Egocentric }           & Body               \\ 
Argus~\cite{duan2024argus}            & BGT60TR13C $\times$ 2 & \textbf{Egocentric}            & Body               \\ \midrule
\textbf{\sysName~(Ours) }                     & IWR6843  & \textbf{Egocentric}            & \textbf{1 or 2 Hands}               \\ \bottomrule
\end{tabular}
\label{tab:existing-work}
\end{table}

\section{Related Work}\label{sec:related_work}

\subsection{Hand Perception with Cameras}\label{sec:visual hci}
In the field of computer vision, a series of methods have been developed for hand pose estimation and gesture recognition, most of which rely on RGB/D cameras that provide rich geometric information~\cite{rezaei2023trihorn, zhang2020differentiable}. Depth cameras perform excellently in handling complex gestures, such as the effective estimation of gesture joints through the combination of the PointNet model ~\cite{ge2018point}, and Zimmermann's proposal of the 3D Transformer network to further improve recognition accuracy ~\cite{zimmermann2017learning}. Besides, ModDrop~\cite{neverova2015moddrop} fuses video stream, depth stream, and audio stream to classify specific gestures. Zhou et al. ~\cite{zhou2021adaptive} proposed an adaptive cross-modal learning method, designing unique modal fusion strategies for different gestures to improve recognition accuracy. TMMF ~\cite{gammulle2021tmmf} focuses on the single-stage recognition of continuous gestures, emphasizing the importance of temporal information in enhancing the continuity of gesture recognition.

However, these vision-based methods inherently depend on capturing detailed visual information of users’ hands and surrounding environments, which raises privacy concerns in sensitive applications and shared spaces. In addition, they may suffer from occlusions in egocentric scenarios, where hands frequently self-occlude or interact with objects (e.g., VR controllers).

\subsection{Hand Perception with mmWave Radars}\label{sec:rf hci}
In the field of radio frequency (RF) sensing, studies have shown that Wi-Fi and customized radars are capable of estimating the poses of multiple people~\cite{wang2019person,zhao2018through,yan2024person,wang2024multi}, but the performance in terms of stability and accuracy under complex environments is still lacking, and often requires additional hardware such as increased antenna count and device numbers to enhance performance. In contrast,  commercial millimeter-wave radar has significant advantages in spatial resolution and penetration ability, providing a reliable and efficient solution for human gesture recognition and pose estimation~\cite{liu2022mtranssee, liu2020real, palipana2021pantomime, santhalingam2020mmasl}. It is noteworthy that recent research has demonstrated the progress of millimeter-wave radar in behavior detection, such as mHomeGes~\cite{liu2020real} and mTransSee~\cite {liu2022mtranssee}, which use millimeter-wave signals to achieve real-time arm gesture recognition and environment-independent gesture recognition, respectively. mmASL~\cite{santhalingam2020mmasl} extracts frequency features from 60GHz millimeter-wave signals and uses a multitask neural network to recognize American Sign Language. Pantomime et al.~\cite{palipana2021pantomime} uses millimeter-wave radar to compute sparse 3D point clouds for gesture recognition in their self-collected dataset.

It can be observed in Table.~\ref{tab:existing-work} that most existing methods rely on radars fixed in a static position, constraining users to operate strictly within the radar’s frontal field of view. To overcome this limitation, mmEgo~\cite{li2023egocentric} and Argus~\cite{duan2024argus} propose mounting radars on the user’s head to enable egocentric body tracking, allowing more natural and unrestricted movements. Building on this design, our work extends egocentric millimeter-wave sensing to fine-grained hand pose estimation and hand gesture recognition, broadening the scope of egocentric radar-based interaction. Moreover, to the best of our knowledge, our system is the first egocentric mmWave-based solution capable of supporting both one-handed and two-handed interactions, significantly expanding the range of applicable interaction scenarios.

\section{Methods}\label{sec:methods}

The technical details are described below.

\subsection{Data Preprocessing}\label{sec:data-preprocessing}

In our setup, we use a TI IWR6843 mmWave radar, a 1.75 W low-power radar, to capture hand movements. mmWave radar transmits and receives frequency modulated continuous wave (FMCW) signals chirp by chirp (64 chirps per frame, 20 frames per seconds), and mixes the received signals with the transmitted signals to obtain an intermediate frequency (IF) signal~\cite{adib20143d, iovescu2017fundamentals,ding2023mi}. 
Due to the sparse point cloud in mmWave radar data, typically consisting of only a few points, hand reflections are inadequately represented, and minor posture changes can cause significant variations, leading to inconsistent representations. To address this issue, we adopted a preprocessing method similar to that in mHomeGes~\cite{liu2020real} to obtain a richer representation of the radar data.

\begin{figure}[h]
    \centering
    \includegraphics[width=1\linewidth]{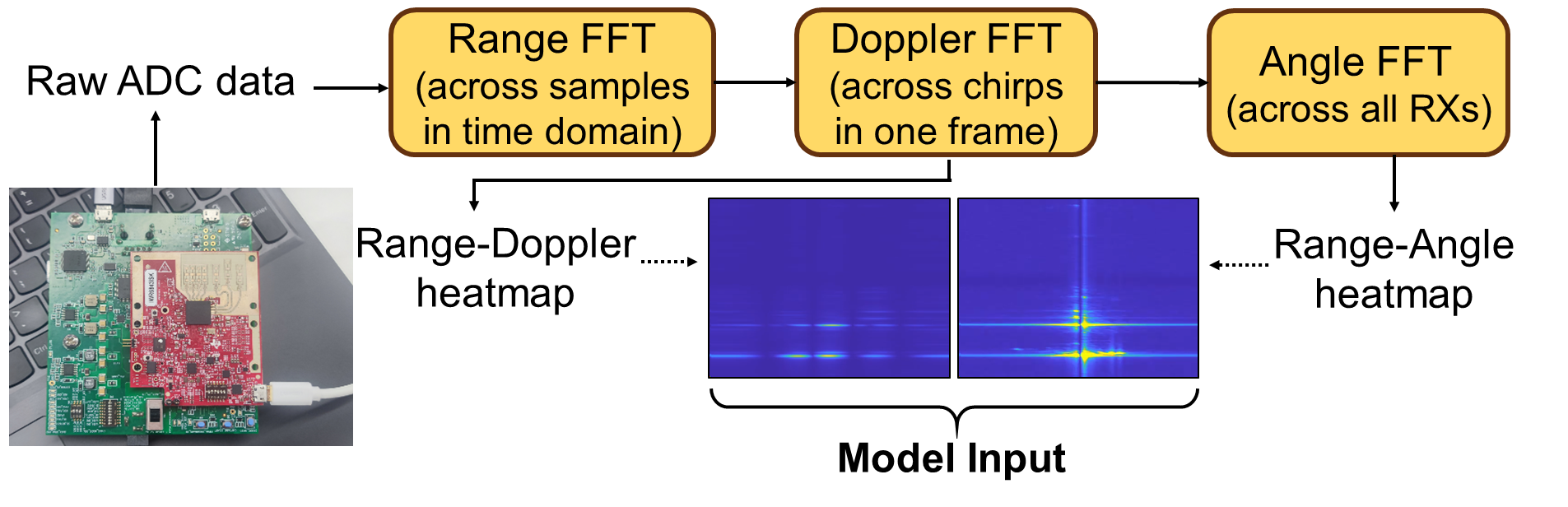}
    \caption{Millimeter-wave radar signal processing involves several Fourier transform steps focusing on the time domain, chirp signals, and the receiving antenna dimension.}
  \label{fig:mmwave_signal}
\end{figure}

$\bullet$ Range FFT. We apply range FFT on every IF chirp, which reveals the frequency difference between the received and transmitted signals, to obtain distance heatmaps of reflection objects. To reduce interference from reflections in the surrounding environment, we retain only the signals within 2 meters after the range FFT, as the distance from the user’s head-mounted mmWave radar to the hands is typically less than 2 meters for most body types.

$\bullet$ Doppler FFT. There exists a phase difference between neighboring chirps caused by the object movements, we use apply Doppler FFT on chirps in one frame along the direction of the phase change caused by the Doppler effect to estimate the object's velocity, the outputs called range-Doppler heatmaps. The velocity information is illustrated at the horizontal axis and the distance information at the vertical axis in Fig.~\ref{fig:mmwave_signal}.

The Range FFT and Doppler FFT can be formulated as follows:
\begin{equation}
  S_{rd} = \mathcal{F}_{chirp}{ (\mathcal{F}_{sampling}{( \text{IF} )})}
\end{equation}
where $S_{rd}$ represents the range-Doppler heatmaps; $\mathcal{F}_{sampling}, \mathcal{F}_{chirp}$ signifies the FFT pertaining to the sampling point dimension and chirp dimension, respectively.

\begin{figure*}[t]
    \centering
    \includegraphics[width=1\linewidth]{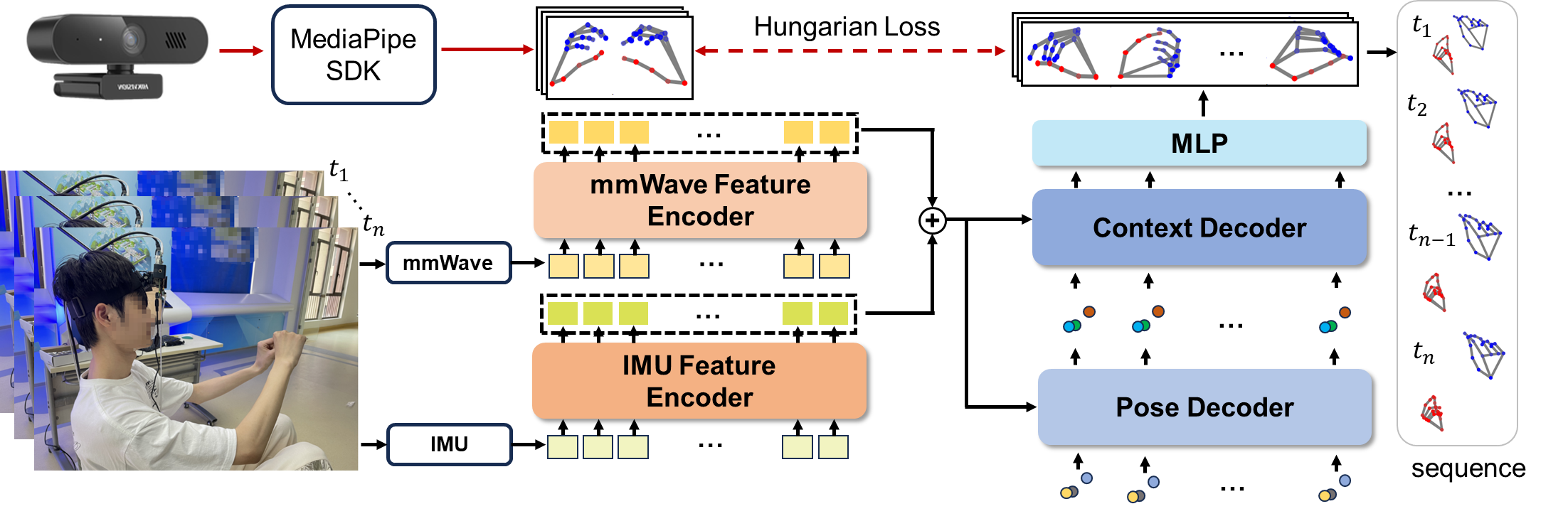}
   \caption{
  \sysName~ takes the head-mounted millimeter-wave radar signals and IMU data to generate hand keypoints. The camera is used solely for label generation during training and is not involved in inference.  }
  \label{fig:hm_network}
\end{figure*}

$\bullet$ Angle FFT. There are phase differences between the antennas caused by the spatial location. We use angle FFT is executed on the range-Doppler heatmaps along the receiving antenna dimension to acquire range-angle heatmaps, which responds to the distance and angle information of the reflection objects. The angle information is illustrated at the horizontal axis and the distance information at the vertical axis in Fig.~\ref{fig:mmwave_signal}. The process can be formulated as follows:
\begin{equation}
  S_{ra} = \mathcal{F}_{rx}{(S_{rd})}
\end{equation}
where $S_{ra}$ denotes the range-angle heatmaps, and $\mathcal{F}_{rx}$ refer to the FFT operation along the receiving antenna dimension.

We concatenate range-Doppler heatmaps and range-angle heatmaps along the vertical axis for the deep network input, as shown in Fig.~\ref{fig:mmwave_signal}.

\subsection{Deep Network Design}

We use TI IWR6843 mmWave radar to capture hand movements. However, the radar signal also contains noise from head movements. To compensate for this, we attach an IMU to the mmWave radar to record its motion. This creates a multimodal fusion problem, and recent research shows that Transformer architectures perform well in multimodal learning~\cite{chen2023immfusion,li2024sensorllm}. Therefore, we use a Transformer-based structure for representation learning.

\begin{figure}[t]
    \centering
    \includegraphics[width=1\linewidth]{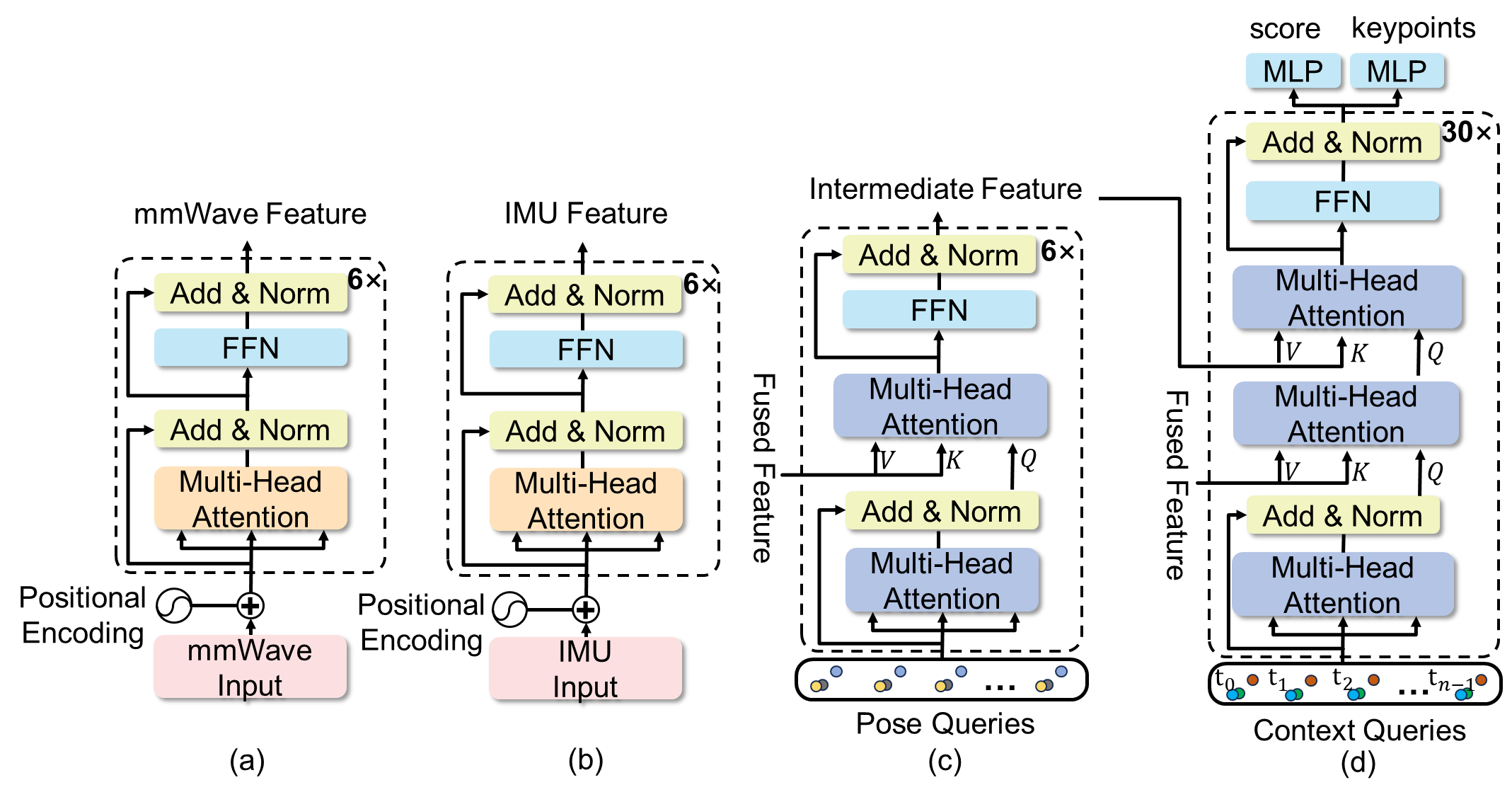}
  \caption{\sysName~ deep network consists of four main novel components. (a) mmWave Radar Encoder, (b) IMU Encoder, (c) Pose Decoder, and (d) Context Decoder. }
  \label{fig:network_structure}
\end{figure}

Another consideration is handling poses involving different numbers of hands: 0, 1, or 2 hands may be present during interactions. Unlike  mmHand~\cite{dong2024mmhand,kong2024mmhand}, which assumes a fixed output of keypoints for one hand, our system must dynamically adapt to variable hand counts. This is similar to the multi-person pose estimation task, where the model must flexibly output coordinates for a variable number of people. To address this, we adapt the Person-in-WiFi 3D framework~\cite{yan2024person}, an end-to-end solution for multi-person pose estimation, to regress hand keypoints for multiple hands. Guided by these insights, we propose the framework shown in Fig.~\ref{fig:hm_network} and the network structure illustrated in Fig.~\ref{fig:network_structure}.

$\bullet$ \textbf{Input and Output.}  We take a 2-second sequence of mmWave radar and IMU signals as input and output the corresponding 3D hand pose sequence. The mmWave input is represented as $x_{mm} \in \mathbb{R}^{30 \times 256 \times 128}$, and the IMU input as $ x_{imu} \in \mathbb{R}^{30 \times 2 \times 3}$, where 30 is the number of sampled frames. The output is a 3D keypoint sequence of shape $\mathbb{R}^{30 \times h \times K \times 3}$, where $h$ is the number of hands, and $K$ is the number of keypoints per hand. To align the modalities temporally, we divide the 2-second window into 30 uniform patches. Each mmWave patch is flattened into a vector of shape $\mathbb{R}^{32,768 \times 1}$ (i.e., $256 \times 128$), and each IMU patch—consisting of one sample across 6 channels (3-axis acceleration and 3-axis angular velocity)—is flattened into a vector of shape $\mathbb{R}^{6 \times 1}$. These patch vectors are then independently fed into an mmWave encoder and an IMU encoder to extract modality-specific temporal features for subsequent fusion.

$\bullet$ \textbf{mmWave Radar Encoder.} This encoder integrates six encoder blocks to process mmWave radar streams for mmWave radar embeddings. Each block comprises a multi-head self-attention module and a Feed-Forward Network (FFN), which are fundamental components of the Transformer architecture~\cite{vaswani2017attention}, shown in Fig.~\ref{fig:network_structure} (a).

$\bullet$ \textbf{IMU Encoder.} This encoder is similar to the mmWave Radar Encoder, which is used for IMU data embeddings. The encoded IMU embeddings are concatenated with the mmWave radar embeddings for further Pose Decoder and Context Decoder, shown in Fig.~\ref{fig:network_structure} (b).

$\bullet$  \textbf{Pose Decoder.} The structure of the decoder layer is the basic module of the Person-in-WiFi 3D~\cite{yan2024person} decoder, and we stacked three such layers. 
As shown in Fig.~\ref{fig:network_structure} (c),  one Pose Decoder block contains one multi-head self-attention module and one multi-head cross-attention module. The multi-head cross-attention module compute attention matrix between Pose Queries ($Q_{pose} = [q_1, q_2, \ldots, q_{C}] \in \mathbb{R}^{C \times D}$) and the fused embeddings ($E_{fused}$) from mmWave Radar Encoder and IMU Encoder. Pose Decoder outputs $X_{pose} \in \mathbb{R}^{C \times D}$, representing features of $C$ hand candidates in $D$ dimension. The process of Pose Decoder can be formulated as follows:
\begin{equation}
  X_{pose} = \text{PoseDecoder}(Q_{pose}, E_{fused})
\end{equation}
where $E_{fused}$ represents the fused features from mmWave radar embeddings and IMU embeddings.

$\bullet$ \textbf{Context Decoder.}  
As shown in Fig.~\ref{fig:network_structure} (d), one Context Decoder block includes one multi-head self-attention module and two multi-head cross-attention modules, and we stack 30 such blocks in the Context Decoder. Context Decoder learns the globally temporal relationships among different frames to refine the hand keypoint estimation. Specifically, Context Decoder utilizes the multi-head cross-attention to compute a cross-attention matrix between the current query and previous pose features from Pose Decoder, iteratively. At last, two Multi-Layer Perceptrons (MLPs) are respectively to generate the keypoints of the hand candidates and the corresponding confidence scores indicating the authenticity of these hand candidates. The process of Context Decoder can be formulated as follows:
\begin{equation}
  H_{set}, H_{score} = \text{ContextDecoder}(Q_{context}, E_{fused}, X_{pose})
\end{equation}
where $Q_{context} \in \mathbb{R}^{C \times D}$ represents $C$ learnable context queries; $H_{set} \in \mathbb{R}^{C \times n \times K \times 3}$ represents K 3D-keypoints of C hand candidates of $n$ frames. $H_{score} \in \mathbb{R}^{C\times n}$ represents confidence scores of these hand candidates in $n$ frames. In our settings, $C=100$, $K=21$, and $n=30$. The Context Decoder consists of 30 Transformer blocks, each responsible for predicting the pose at a single time step. Each block takes as input the current frame along with the refined representation from the previous frame, enabling cross-frame temporal modeling through cross-attention.

This dual-decoder design explicitly decomposes the egocentric hand tracking challenge: the Pose Decoder  specializes in spatial joint localization by resolving per-frame ambiguities through cross-attention between pose queries $Q_{pose}$ and sensor embeddings $E_{fused}$; conversely, the Context Decoder captures temporal gesture kinematics via iterative feature refinement across frames, enabling robust recognition of dynamic motions. This separation aligns with the observation that spatial precision and temporal coherence require specialized processing under head-motion interference.

$\bullet$ \textbf{Loss Function.} We implement a Hungarian Matching algorithm~\cite{kuhn1955hungarian} to ensure a unique prediction for the ground-truth 3D hand keypoints of each hand candidate.
\begin{equation}
  L_{kpt} = \text{HungarianMatch}(H_{set}, H_{gt})
\end{equation}
where $H_{gt}$ denotes the ground-truth hand keypoint coordinates. $H_{set}$ is selected up to two hand candidates based on $H_{score}$.

$\bullet$ \textbf{Implementation Details.}
We employed the Adam optimizer~\cite{kingma2014adam} to train a deep network on an NVIDIA 3090 GPU. The batch size was 32.  The training was carried out for 500 epochs. The initial learning rate was 0.001 and was halved every 100 epochs. At the start of each epoch, the training samples were randomly shuffled. 

The resulting model contains 55.42M parameters, with an inference GPU memory of 657 MB and an end-to-end inference time of 42.03ms for 30 frames on the RTX 3090 GPU. These results demonstrate that our framework already achieves real-time performance.

$\bullet$ \textbf{Downstream gesture recognition task.} The output of \sysName~ can be seamlessly applied to downstream tasks such as gesture recognition. Given a sequence of predicted hand poses $H_{set}$, we feed it into a gesture classification network (e.g., a GNN or ResNet) to identify the performed gesture. To ensure a consistent input size for the recognition model, we handle single- and dual-hand scenarios differently: if two hands are present, their keypoints are concatenated directly; if only one hand is detected, its keypoints are duplicated before concatenation. This design allows the recognition network to operate on a fixed-size input regardless of the number of hands.

\section{Experiments}\label{sec:experiments}

\subsection{Data Acquisition}
Data was collected under the approval of the IRB.

\begin{figure}[h]
  \centering
  \includegraphics[width=1\linewidth]{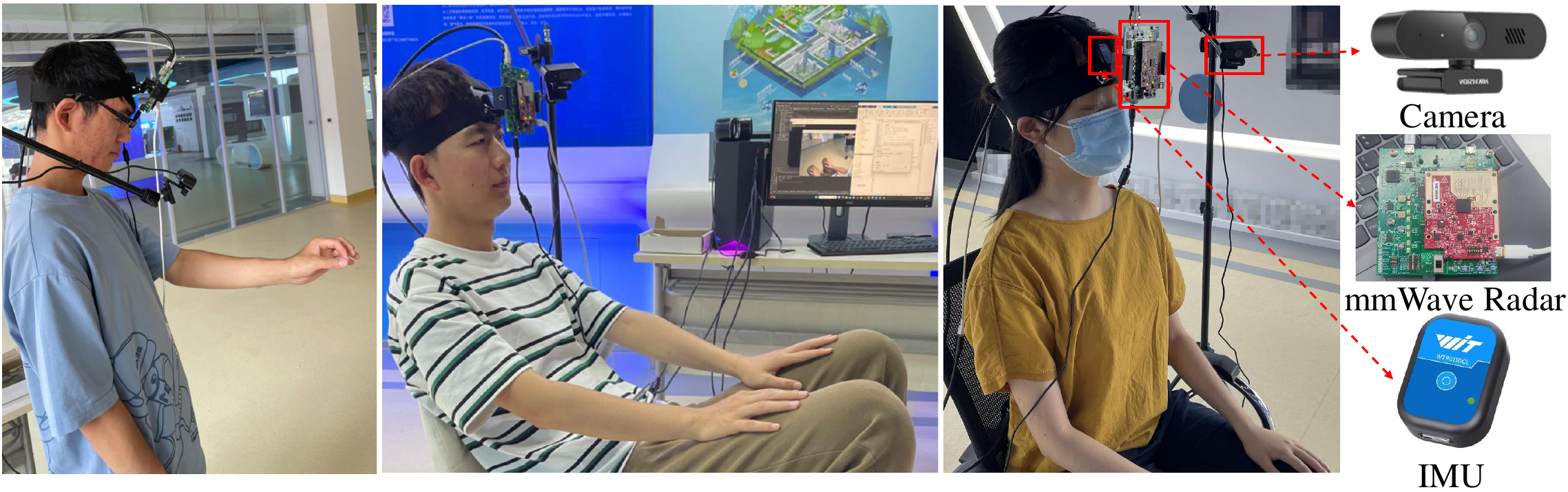}
  \caption{Hardware setup. The data collection hardware consists of a millimeter-wave radar, an IMU, and a camera, including three postures in three scenes, i.e., standing at scene \#3, lying at scene \#2, and sitting at scene \#1, respectively. }
  \label{fig:hardware}
\end{figure}

\textbf{(1) Hardware Configurations.}
The data collection hardware consists of a millimeter-wave radar, an IMU, and a camera, shown in Fig.~\ref{fig:hardware}. The millimeter-wave radar and IMU device are mounted on a plastic bracket and secured to the head using an adjustable strap. A tripod is used to position the camera for hand motion capture and hand pose labeling. All three devices are controlled and synchronized via a desktop computer, ensuring precise temporal synchronization during data recording.

$\bullet$ Millimeter-wave radar.  One TI IWR6843ISK radar generates frequency modulated continuous wave (FMCW) signals at frequencies of 60GHz-64GHz. Its power consumption is only 1.75W, slightly lower than typical Wi-Fi devices, making it safe for human exposure. The radar has three transmitting antennas and four receiving antennas. The transmit parameters are set to 20 frames per second with 64 chirps per frame. The analog-to-digital converter (ADC) sampling rate is 256, and the raw radar data is obtained by mounting the radar on a TI DCA1000EVM board. The raw radar data is $\in 20t \times 3 \times 4 \times 256 \times 64$, where $t$ is the time in seconds. The range-Doppler and range-angle heatmaps are obtained after the preprocessing process described in Sec.~\ref{sec:data-preprocessing}, both of which are $\in 20t \times 256 \times 64$. These heatmaps are then combined along the distance dimension to obtain model inputs $ \in 20t \times 256 \times 128$.

$\bullet$ IMU. The IMU sensor records the acceleration and angular velocity data in the X, Y, and Z axes at a rate of 20 samples per second. The recorded samples are $ \in 20t \times 2\times 3$, where $t$ is the time in seconds.

$\bullet$ Camera. A Hikvision camera captures the RGB videos at 20 frames per second in $1080\times 720$ resolution. 

\textbf{(2) Dataset diversity.} We recruited 10 volunteers, whose heights ranged from 161 to 182 centimeters and weights from 44 to 73 kilograms. The experiment was conducted across three different environments and under three distinct postures: sitting, lying, and standing. In each condition, the participants performed eight types of human-computer interaction gestures: click, swipe left, swipe right, swipe up, swipe down, swipe to the bottom-right, zoom in, and zoom out. Among these, the last two are two-handed gestures, while the rest are performed with a single hand. The selection of gestures and postures was inspired by the Apple Vision Pro, which employs similar interactions—such as confirming selections or scrolling through content—for a broad range of VR functionalities. In every scene and posture, each volunteer was asked to perform each gesture 20 times, with each gesture completed within a 2-second duration.

Among the 10 volunteers, two are left-handed, and in performing single-handed gestures, they were instructed to perform gestures using their left and right hands, respectively. The remaining eight volunteers are right-handed; six of them performed gestures with their dominant right hand, while the other two were assigned to use their non-dominant left hand.
Consequently, there are $2\times2+6+2=12$ different settings for the left and right hand executing gestures in total.  Table.~\ref{tab:volunteer-diversity} summarizes this experimental configuration. This setup has been designed for scenarios where a user's preferred hand is occupied, ensuring that the system can still recognize and respond to gestures executed by the non-preferred hand.

\begin{table}[t]
  \footnotesize
    \setlength{\tabcolsep}{3pt} 
  \centering
    \caption{We recruited 10 volunteers to perform VR interaction gestures. They were asked to execute  gestures by their preferred hand or non-preferred hand. Pre.Hand for Preferred hand; Exe.Hand for Executing Hand; H. for Height; W. for Weight.}
  \begin{tabular}{cccccccc}
  \toprule
  No.  & ID & Pre.Hand & Exe.Hand & Gender &H.(cm) &W.(kg) &BMI\\ 
  \midrule
  1 & 1 & Left & Right & Female &173 &58 & 19.38 \\
  2 & 1 & Left & Left  & Female &173 &58 & 19.38 \\
  3 & 2 & Left & Right  & Male &175 &73 & 23.84 \\
  4 & 2 & Left & Left & Male &175 &73 & 23.84 \\
  5 & 3 & Right & Left & Male &172 &62 & 20.96\\
  6 & 4 & Right & Right  & Female &165 &45 & 16.53 \\
  7 & 5 & Right & Right & Male &182 &67 & 20.23 \\
  8 & 6 & Right & Right & Male &174 &68 & 22.46 \\
  9 & 7 & Right & Right & Male &161 &44 & 16.97 \\
  10 & 8 & Right & Right & Male &178 &65 & 20.52 \\
  11 & 9 & Right & Right & Male &175 &50 & 16.33 \\
  12 & 10 & Right & Left  & Female &163 &48 & 18.07\\
  \bottomrule
  \end{tabular}
  \label{tab:volunteer-diversity}
\end{table}

\textbf{(3) Dataset statistics.}\label{sec:dataset statistics}
The dataset comprises a total of 5,760 gesture samples, collected across 3 scenes, 12 hand-executing settings, 8 gesture types, and 20 repetitions per gesture (3×12×8×20). Each sample includes synchronized data from millimeter-wave radar, IMUs, and video recordings. For data partitioning, we uniformly select the 5th, 10th, 15th, and 20th repetitions for testing, and use the remaining 16 repetitions for training. To provide supervision for deep network training and evaluation, we employ the Google MediaPipe Hand Landmark SDK to extract 3D hand keypoints from the video data. However, we observe that the SDK occasionally fails to detect keypoints, particularly when the hand moves rapidly. If more than 40\% of frames in a video are missing keypoints, the corresponding sample is discarded. Otherwise, the extracted keypoints are temporally \textit{\textbf{downsampled or upsampled to 30 frames per gesture sample}} to maintain consistency. Although MediaPipe’s 3D keypoints may not offer high-precision tracking, this automated approach significantly reduces the annotation burden. After filtering out invalid samples, the final dataset contains 5,206 gesture instances. Detailed statistics are presented in Table~\ref{tab:dataset_statistics_hm}.

\begin{table}[t]
    \caption{Dataset samples. Volunteers executed the gestures with postures of sitting in scene \#1, lying in in scene \#2, and standing at scene \#3, respectively.}
  \setlength{\tabcolsep}{3pt} 
    \begin{tabular}{lcccc}
    \toprule
     &  sitting in s\#1 & lying in s\#2  & standing in s\#3 & all                   \\
    \midrule
    \#training  & 1384 & 1384 & 1384 & 4152           \\
    \#test  & 350  & 351  & 353  & 1054            \\
    \bottomrule
    \end{tabular}
    \label{tab:dataset_statistics_hm}
\end{table}

\subsection{Evaluation Metrics}

\textbf{(1) Mean Per Joint Position Error(MPJPE).}
The L2 norm is computed between the predicted and true 3D coordinates of the hand joints, as shown in Equation \ref{eq:pjpe}. This computation leads to the Joint Position Error.
\begin{equation}\label{eq:pjpe}
    \operatorname{PJPE}(\mathrm{k})=\frac{1}{F} \sum_{f=1}^F |p r e(f, k)-g t(f, k)|_2
\end{equation}
\vspace{-7pt}
\begin{equation}\label{eq:mpjpe}
    \operatorname{MPJPE}=\frac{1}{K} \sum_{k=1}^K E_{P J P E}(k)
\end{equation}
$\operatorname{PJPE}(\mathrm{k})$ is the PJPE for the k-th joint. The MPJPE is the average of all $\operatorname{PJPE}(\mathrm{k})$.

\textbf{(2)}  We also adopt gesture recognition Accuracy, Precision, Recall, and F1-score for the evaluation.

\begin{table*}[t]
  \caption{ \sysName~outperforms one-stage methods such as mGesNet~\cite{liu2020real}, mSeeNet~\cite{liu2022mtranssee}, and  mmGesture~\cite{yan2023mmgesture}, by a large margin. MPJPE is measured in millimeters(mm). mm4Arm~\cite{liu2022leveraging} is a single-hand pose estimation method, trained and evaluated solely on single-hand data. Compared to simultaneous detection of both single-hand and two-hand poses, this is a considerably simpler task. Nevertheless, our method significantly outperforms mm4Arm despite addressing the more challenging setting.
  }
  \centering
  \begin{tabular}{lcccccc}
  \toprule
  Metrics  & MPJPE$\downarrow$ & Acc(\%) & Precision & Recall & F1\\ 
  \midrule
\rowcolor{red!10} \textbf{\sysName~(mmWave+IMU)} & \textbf{72.73mm} & / & / & / & /\\ 

\rowcolor{red!10} \sysName~(mmWave)  &  96.42mm & / & / & / & / \\
\rowcolor{red!10} \sysName~(mmWave+IMU, w/o Context Decoder) & \textbf{109.02mm} & / & / & / & /\\ 

\rowcolor{red!10} mm4Arm~\cite{liu2022leveraging}$^*$~(mmWave+IMU) & \textbf{165.19mm} & / & / & / & /\\ 
\rowcolor{blue!10} two-stage (\textbf{\sysName+ResNet50~\cite{he2016deep}}) & / & \textbf{90.80} & \textbf{93.34} & \textbf{93.20} & \textbf{93.19}\\ 
\rowcolor{blue!10}  two-stage (\sysName+LSTM)        & / & 89.26 & 91.56 & 91.48 & 91.46\\ 
\rowcolor{blue!10} two-stage (\sysName+GCN)         & /& 85.36 & 87.16 & 87.12 & 87.09 \\ 
\rowcolor{blue!10} two-stage (\sysName+ViT~\cite{dosovitskiy2020image})         & / & 77.95 & 81.39 & 80.19 & 79.28  \\

\rowcolor{green!10} one-stage (mSeeNet~\cite{liu2022mtranssee})   & / & 84.60 & 86.73 & 86.59 & 86.60\\ 
\rowcolor{green!10} one-stage (mGesNet~\cite{liu2020real})   & / & 81.34 & 81.98 & 81.56 & 81.64\\
\rowcolor{green!10} one-stage (mmGesture~\cite{yan2023mmgesture}) & / & 68.66 & 70.01 & 68.83 & 68.81\\ 
\rowcolor{green!10} one-stage (LSTM)      & / & 67.99 & 70.58 & 68.35 & 67.80 \\
\rowcolor{green!10} one-stage (ResNet50~\cite{he2016deep})       & / & 60.09 & 61.79 & 61.19 & 60.64\\
\rowcolor{green!10} one-stage (ViT~\cite{dosovitskiy2020image})       & / & 59.66 & 67.12 & 59.77 & 60.75 \\
\bottomrule
  \end{tabular}
  \label{tab:backbone_hm}
\end{table*}

\subsection{Results}\label{sec:main_results}

\textbf{(1) Overall performance.} 

$\bullet$  \textbf{IMU works.}  Table~\ref{tab:backbone_hm} presents a comprehensive evaluation of \sysName's performance in both hand pose estimation and downstream gesture recognition, following the data partitioning strategy detailed in Table~\ref{tab:dataset_statistics_hm}. Our system achieves a Mean Per Joint Position Error (MPJPE) of 72.73mm, demonstrating a 24.6\% improvement over mmWave-only approaches (96.42mm MPJPE). 
This significant enhancement not only confirms the presence of head motion in our data collection process but also validates the effectiveness of our IMU signal fusion strategy for egocentric hand pose estimation.

$\bullet$   \textbf{Context Decoder works.}  To further assess the contribution of the Context Decoder, we conducted a controlled experiment by training the model with only the Pose Decoder. In this configuration, the system achieved an MPJPE of 109.92mm. These results confirm that the Context Decoder significantly enhances performance by modeling sequential context across frames.

$\bullet$ \textbf{SOTA Pose Estimation.}  Besides, due to the current limitations of mmHand and mm4Arm—namely, their support for single-hand estimation only and their closed-source implementations~\cite{dong2024mmhand,kong2024mmhand,liu2022leveraging}—we independently reimplemented mm4Arm~\cite{liu2022leveraging} and trained it exclusively on single-hand data for hand pose estimation, achieving an MPJPE of 165.19mm. This single-hand setting is inherently simpler than our scenario, which simultaneously estimates both single- and two-hand poses. Nevertheless, our method significantly outperforms mm4Arm despite addressing this more challenging task.

$\bullet$  \textbf{SOTA Gesture Recognition.} Furthermore, we use the hand poses estimated by \sysName's intermediate features and feed them into simple classification models such as ResNet~\cite{he2016deep}, LSTM, GCN, and ViT~\cite{dosovitskiy2020image} for gesture recognition, as highlighted in blue in Table\ref{tab:backbone_hm}. All models achieve strong performance under this two-stage pipeline. For instance, using ResNet-50 yields a gesture recognition accuracy of 90.80\% and an F1-score of 93.19. These results are 20–30 percentage points higher than directly applying LSTM, ResNet-50, or ViT to raw mmWave and IMU data. Moreover, \sysName~ also outperforms carefully designed state-of-the-art methods for mmWave gesture recognition, such as mGesNet~\cite{liu2020real}, mSeeNet~\cite{liu2022mtranssee}, and  mmGesture~\cite{yan2023mmgesture}. This finding suggests that future research could benefit from incorporating hand pose estimation as an intermediate representation to enhance gesture recognition performance.

\begin{table}[t]
  \centering
    \caption{Accuracy, Precision, Recall and F1 for each gesture.}
  \begin{tabular}{lcccc}
  \toprule
  Gesture  & Acc(\%)  & Precision  & Recall  & F1 \\
  \midrule
  Click  & 99.21 & 96.15 & 99.21 & 97.66  \\
  Swipe leftward  & 97.78 & 96.35 & 97.78 & 97.06  \\ 
  Swipe rightward & 97.06 & 96.35 & 97.06 & 96.70  \\ 
  Swipe upward   & 88.43 & 97.27 & 88.43 & 92.64 \\ 
  Swipe downward & 94.74 & 95.45 & 94.74 & 95.09 \\ 
  Swipe to bottom right    & 100.00 & 97.10 & 100.00 & 98.53  \\  
  Zoom in   & 79.51 & 87.39 & 79.51 & 83.26  \\ 
  Zoom out   & 88.89 & 80.62 & 88.89 & 84.55  \\ 
  mean      & 90.80  & 93.34  &  93.20 &  93.19 \\ 
  \bottomrule
  \end{tabular}
  \label{tab:gesture}
\end{table}

\textbf{(2) PJPEs for fingers.} In addition, the PJPEs of the thumb, index, middle, ring, and little fingers are 60.18, 70.10, 76.42, 84.44, and 91.79, respectively. The thumb and index fingers exhibit the smallest prediction errors, while the ring and little fingers show the largest. This is because, from the perspective of a head-mounted mmWave radar, the thumb and index fingers are more prominently exposed, whereas the other fingers are often occluded by them, making accurate estimation more challenging. Nevertheless, thanks to the penetration capability of mmWave signals~\cite{li2023egocentric}, our model is still able to achieve reasonable estimation performance for these occluded fingers.

\begin{figure*}[t]
  \centering
  \includegraphics[width=1\linewidth]{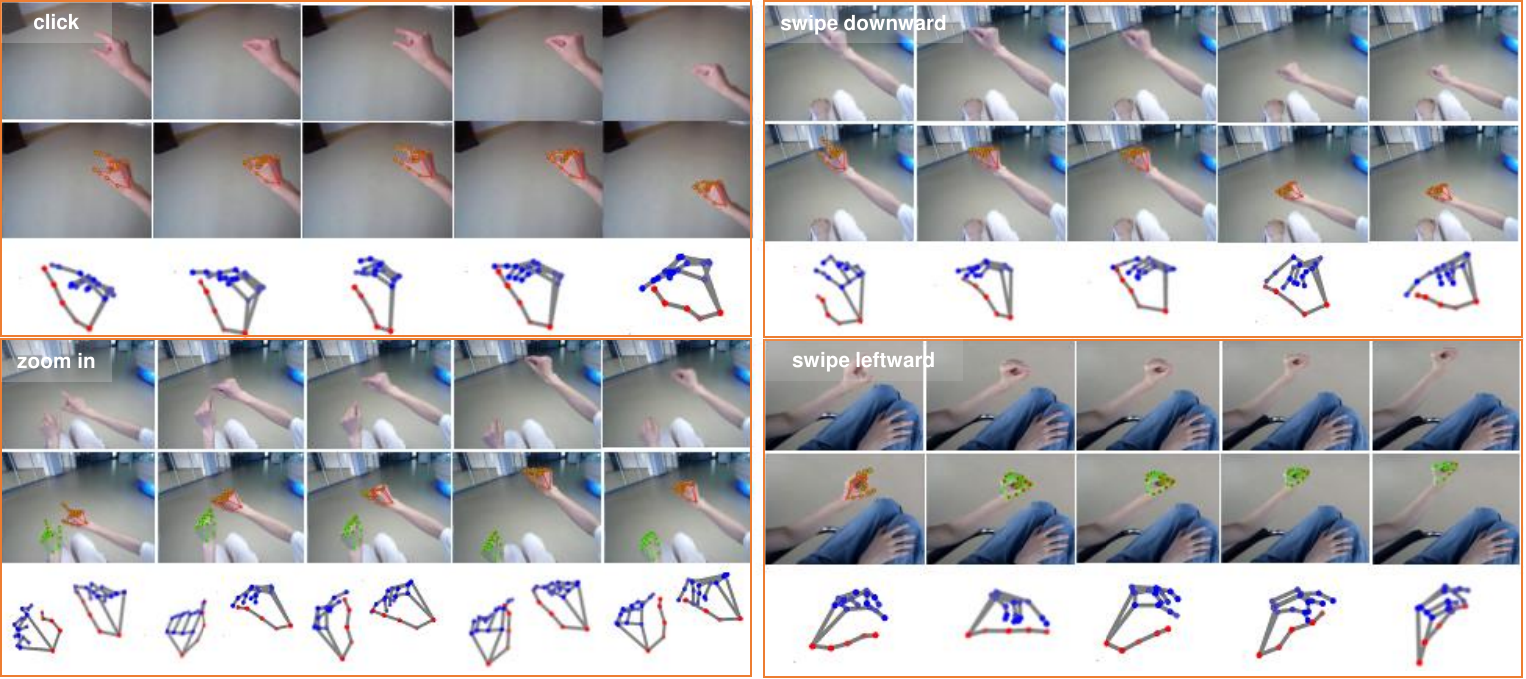}
  \caption{\sysName~ hand pose estimation examples, showing click, swipe downward, zoom in, and swipe leftward, respectively. (Images here are only used for visualization).}
  \label{fig:pose_example}
\end{figure*} 

\textbf{(3) Visualization.} Fig.~\ref{fig:pose_example} illustrates the visualization of hand keypoint predictions by the \sysName~ for various gestures, showcasing different volunteers performing gestures such as click, swipe downward, zoom in, and swipe leftward. Each set of gestures is depicted in three rows: the first row presents the video frames captured during the gestures, the second row shows the hand keypoints derived from the video frames using the Google SDK, and the third row displays the predicted hand keypoints output by \sysName. It can be seen that \sysName~ can automatically detect the number of hands, whether they are left or right hands, as well as the movement of hand joints throughout the gesture executing process, which effectively alleviates the difficulty of subsequent gesture recognition.

\begin{figure}[t]
  \centering
  \includegraphics[width=1\linewidth]{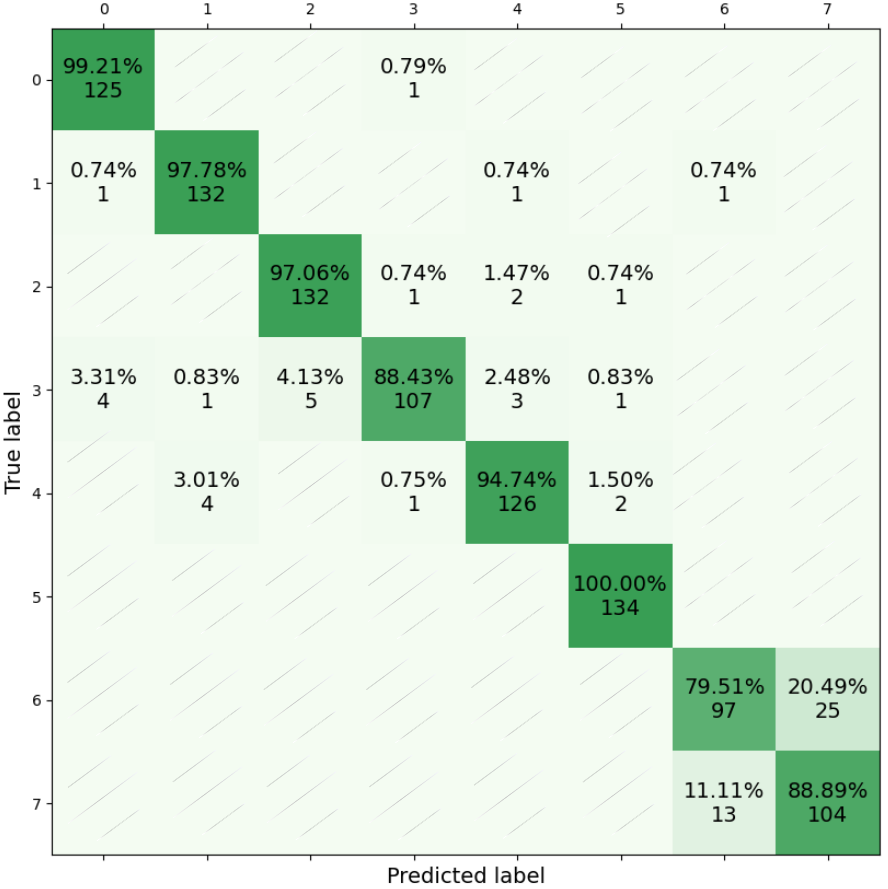}
  \caption{The confusion matrix of gesture recognition.}
  \label{fig:cm_hm}
\end{figure} 

\textbf{(4) Downstream gesture recognition.} Fig.~\ref{fig:cm_hm} presents the confusion matrix for recognizing eight human-computer interaction gestures. Each cell in the matrix indicates both the percentage and the absolute number of samples classified into each category. Table~\ref{tab:gesture} further reports the Accuracy, Precision, Recall, and F1 Score for each gesture. We observe that single-hand gestures generally achieve high accuracy, often exceeding 90\%, while two-hand gestures exhibit slightly lower performance, though still above 80\%. The most prominent recognition errors occur between the two-hand gestures zoom in and zoom out, which are frequently misclassified with each other. This confusion is largely attributed to the lack of strict instructions given to participants during data collection regarding the specific diagonal direction of movement for these gestures, resulting in greater intra-class variability. Additionally, we observe that the MPJPE for zoom in and zoom out is notably higher compared to single-hand gestures, contributing to reduced accuracy in the downstream gesture classification task.

\subsection{Few-shot Gesture Recognition}

We further evaluate the generalization ability of \sysName~for gesture recognition under few-shot fine-tuning in challenging real-world scenarios, including cross-person, cross-posture, and cross-hand settings. Please note that the primary goal of this evaluation is to demonstrate the baseline generalization capability of our framework. Achieving stronger generalization performance would require incorporating additional techniques such as domain adaptation and domain generalization. To facilitate future research in this direction, we release our dataset publicly to support further exploration of these advanced generalization methods.

\textbf{(1) Cross-person.} We adopt two cross-person validation strategies to evaluate the gesture recognition performance of \sysName~ on unseen users. (1) In the first strategy, we use data from No.1 to No.8 (as listed in Table.~\ref{tab:volunteer-diversity}) for training, and test the model on data from No.9 to No.12. (2) In the second strategy, we perform leave-one-subject-out evaluation: in each iteration, data from one row in Table.~\ref{tab:volunteer-diversity} is used for testing, while data from the remaining rows is used for training. This process is repeated for all 12 rows, and the average performance is reported. As shown in Table~\ref{tab:cross-people-hm}, \sysName~demonstrates a certain degree of generalization ability to unseen users, achieving gesture recognition accuracies of 63.11\% and 61.96\% under the two strategies, respectively. Furthermore, the accuracy improves significantly—by 10\% to 20\%—when one-shot or two-shot fine-tuning is applied, using only one or two labeled samples from the target user to adapt the \sysName.

\begin{table}[t]
    \caption{Gesture recognition tests on unseen volunteers. \sysName~ maintains a certain level of recognition capability for unseern persons, and can be improved by 10\%-20\% with one-shot or two-shot fine-tuning.}
  \footnotesize
  \setlength{\tabcolsep}{2pt}
  \centering
  \begin{tabular}{cccccc}
    \toprule
    \multicolumn{1}{c}{\#Training; \#Test}    & \multicolumn{1}{c}{\#-Shot}          & Acc(\%) & Precision & Recall & F1\\ 
    \midrule
         \multicolumn{1}{c}{} & zero  & 63.11  & 64.42 & 63.17 & 62.74 \\ 
          \multicolumn{1}{c}{No.1-8; No.9-12 (Table.~\ref{tab:volunteer-diversity})} & one & 73.99  & 74.50 & 73.60 & 73.57\\ 
     \multicolumn{1}{c}{} & two  & 83.97  & 84.13 & 83.88 & 83.87 \\ 
     \midrule
          \multicolumn{1}{c}{} & zero  & 61.96 & 68.16 & 65.54 & 64.68 \\ 
          \multicolumn{1}{c}{leave-one-person-out} & one  & 69.37 & 74.97 & 74.41 & 73.85 \\ 
          \multicolumn{1}{c}{} & two  & 74.65 & 77.86 & 76.51 & 75.09 \\ 
  \bottomrule
  \end{tabular}
  \label{tab:cross-people-hm}
\end{table}

\begin{figure}[t]
  \centering
  \includegraphics[width=1\linewidth]{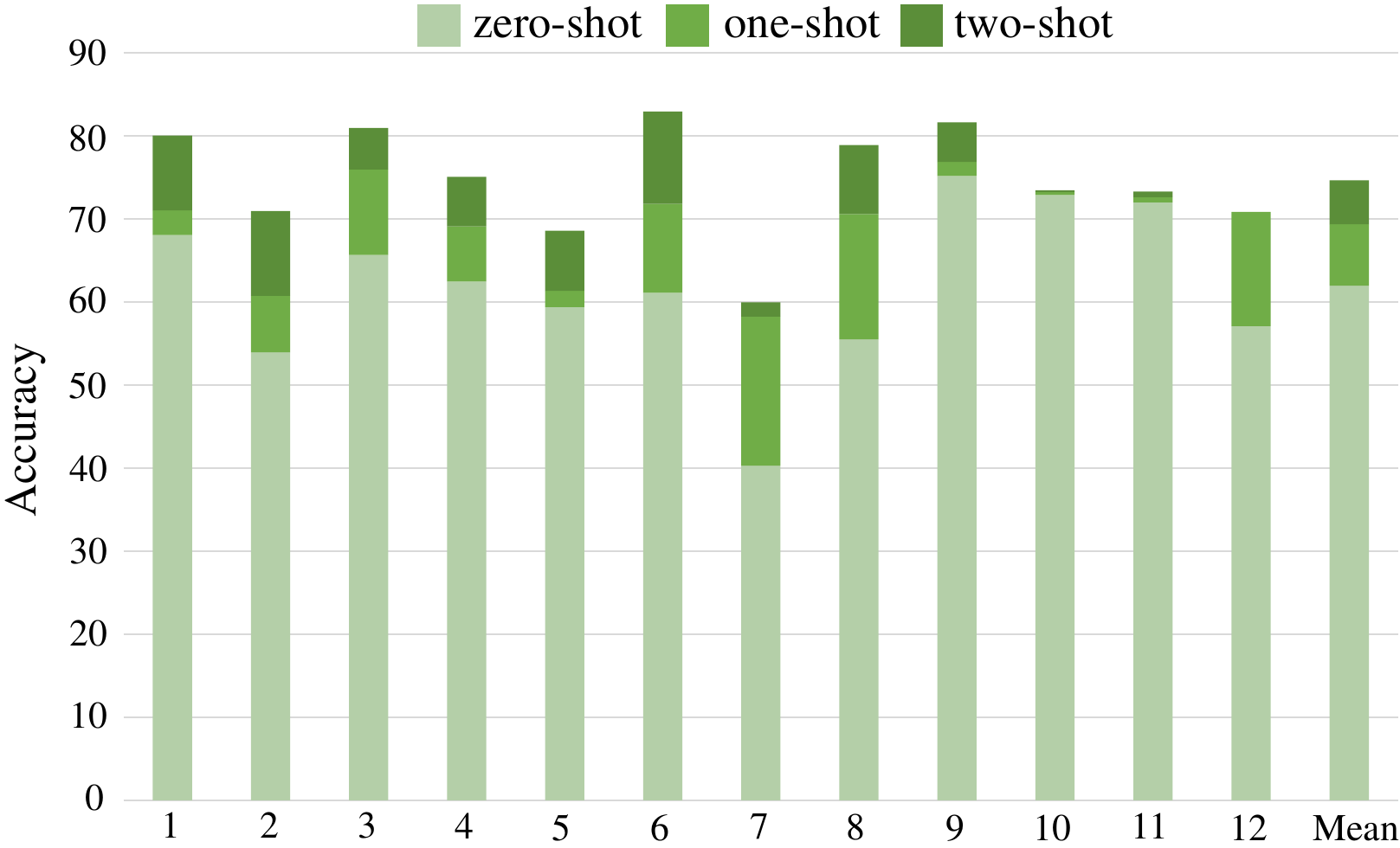}
  \caption{Gesture recognition accuracy in leave-one-person-out setting.}
  \label{fig:cross-person-hm}
\end{figure}

Fig.~\ref{fig:cross-person-hm} demonstrates the leave-one-person-out results. It also shows that \sysName~ maintains a certain level of recognition capability and can be improved further by few-shot fine-tuning. There is one exception observed with No. 7, which exhibits a significantly lower accuracy rate. Upon reviewing the video footage, we find that this particular volunteer performs the gestures with smaller and more casual movements, deviating considerably from the patterns exhibited by others. Consequently, during actual system deployment, what we need is to implement a \textit{\textbf{pre-use calibration protocol}}, requiring users to calibrate each gesture twice or more before commencing usage.

\textbf{(2) Cross-posture and Cross-scene.} We use data from two postures (or scene) in Table.~\ref{tab:dataset_statistics_hm} to train \sysName~and tests the trained \sysName~on data of the last postures (or scene). Similar to the cross-person performance,  \sysName~ maintains a certain level of recognition capability for unseen postures/scenes, achieving a gesture recognition accuracy of 54.35\%, 51.42\%, and 51.41\% in three cross-posture evaluation, respectively. Moreover, the gesture recognition accuracy significantly improves by 20\% with one-shot or two-shot fine-tuning. This experiment further substantiates the efficiency of pre-use calibration strategy discussed in the above cross-person evaluation.

\begin{table}[thbp]
  \caption{Gesture recognition tests on unseen posture/scene. \sysName~ maintains a certain level of recognition capability for unseen posture/scene, and can be improved by 20\% with one-shot or two-shot fine-tuning.}
  \footnotesize
      \setlength{\tabcolsep}{2pt} 
  \centering
  \begin{tabular}{ccc}
  \toprule
  \#Training; \#Test & \#-Shot & Acc(\%) \\ 
  \midrule
   sitting in scene \#1 (dataset split in Table~\ref{tab:dataset_statistics_hm}) & / & 87.60   \\ 
   lying in scene \#2 (dataset split in Table~\ref{tab:dataset_statistics_hm}) & / & 87.96   \\ 
  standing in scene \#3 (dataset split in Table~\ref{tab:dataset_statistics_hm}) & / & 95.92   \\
  \hline     
  
  & zero  & 54.35   \\ 
  scene \#2 (lying) and \#3 (standing); scene \#1 (sitting) & one & 67.58   \\ 
  & two  & 74.97   \\ 
  
  \cline{2-3}
  
  & zero  & 51.42   \\ 
  scene \#1 (sitting) and \#3 (standing); scene \#2 (lying) & one & 70.33   \\ 
  & two  & 76.97   \\
  \cline{2-3}
  
  & zero  & 51.41   \\ 
  scene \#1 (sitting) and \#2 (lying); scene \#3 (standing) & one & 60.42   \\ 
  & two  & 74.51   \\
  \bottomrule
  \end{tabular}
  \label{tab:cross-scene-hm}
\end{table}

\begin{table}[t]
  \caption{Gesture recognition tests on unseen hands. \sysName~ maintains a weak level of recognition capability for unseern hands, and can be largely improved by 30\%-50\% with one-shot or two-shot fine-tuning.}
  \footnotesize
    \setlength{\tabcolsep}{2pt} 
  \centering
  \begin{tabular}{ccccc}
  \toprule
  Data splitting  strategy & \multicolumn{1}{c}{\#Training Hand}  & \multicolumn{1}{c}{\#Test Hand}      & \multicolumn{1}{c}{\#-Shot}  & Acc(\%) \\ 
  \midrule
  \multirow{2}{*}{Table~\ref{tab:dataset_statistics_hm}}   & \multicolumn{2}{c}{accuracy of left hand } & / & 90.96   \\ 
        & \multicolumn{2}{c}{accuracy of right hand } & / & 93.04   \\ 
  \hline     
  
  \multirow{6}{*}{Cross-hand evaluation} & \multicolumn{1}{c}{} & \multicolumn{1}{c}{}   & zero  & 28.25   \\ 
   & \multirow{1}{*}{Right} & \multirow{1}{*}{Left}     & one & 57.80   \\ 
    & \multicolumn{1}{c}{}  & \multicolumn{1}{c}{}    & two  & 73.53   \\ 
  \cline{2-5}
  
  & \multicolumn{1}{c}{} & \multicolumn{1}{c}{}   & zero  & 24.95   \\ 
  \multirow{1}{*}{} & \multirow{1}{*}{Left} & \multirow{1}{*}{Right}     & one & 58.92   \\ 
  & \multicolumn{1}{c}{}  & \multicolumn{1}{c}{}    & two  & 77.18   \\
  \bottomrule
  \end{tabular}
  \label{tab:cross-hand-hm}
\end{table}

\textbf{(3) Cross-hand.} Table.~\ref{tab:cross-hand-hm} shows that \sysName~ achieves that the accuracy for each hand exceeds 90\% if following the data splitting strategy (Table~\ref{tab:dataset_statistics_hm}). We also conduct cross-hand evaluation, training \sysName~ with data of right/left executing hand in Table.~\ref{tab:volunteer-diversity}, and test the trained \sysName~ with data of left/right executing hand, respectively. Note that this training strategy also encompasses cross-person challenges. For example, as shown in Table.~\ref{tab:volunteer-diversity}, instances from No.6 to No.11 introduce cross-person challenges when the left-hand execution data is used for training and the right-hand execution data is employed for testing. Thus, this is a highly challenging data splitting scheme. However, upon fine-tuning with just one to two samples, gesture recognition accuracy sees a remarkable improvement of 30-50\%, up to 73.53\% and 77.18\%. This indicates that during the training of \sysName, we should simultaneously provide samples from both the left and right hands.

\section{Conclusion}

This paper presents \sysName, a proof-of-concept system for egocentric hand pose estimation and gesture recognition, leveraging head-mounted millimeter-wave radar and IMUs. This configuration facilitates user mobility while offering enhanced personal privacy compared to traditional camera-based solutions. We conduct extensive experiments to evaluate \sysName, demonstrating the importance of including samples from both hands during training to improve cross-hand generalization. Furthermore, we find that performing at least two pre-use calibration sessions is critical to achieving robust performance across diverse users, environments, and hand postures.

While \sysName~ shows promising results, the current prototype is somewhat cumbersome, especially due to the use of the TI mmWave radar module. Future versions could benefit from more compact hardware, such as Google’s Soli radar (9mm$\times$9mm)~\cite{lien2016soli}. Additionally, mmWave radar has the potential to reveal health-related metrics such as respiration and heart rate~\cite{yang2016monitoring, alizadeh2019remote}, which raises privacy concerns. These can be mitigated by appropriately limiting the radar's range and field of view.

Moreover, although \sysName~ achieves promising accuracy in gesture recognition, there remains room for improvement in the precision of hand pose estimation and its generalization under cross-domain scenarios. To facilitate further research and development, we have publicly released our dataset, enabling the community to explore, benchmark, and improve upon our work.

\IEEEpeerreviewmaketitle

\ifCLASSOPTIONcaptionsoff
  \newpage
\fi



\bibliographystyle{IEEEtran}

\bibliography{IEEEexample}

\end{document}